\documentclass[sigconf,natbib=false]{acmart}
\AtBeginDocument{%
  }

\setcopyright{acmlicensed}
\copyrightyear{2026}
\acmYear{2026}
\acmDOI{XXXXXXX.XXXXXXX}
\acmConference[SAC'26]{The 41st ACM/SIGAPP Symposium on Applied Computing}{March 23--27, 2026}{Thessaloniki, Greece}
\acmISBN{979-X-XXXX-XXXX-X/26/03}

\RequirePackage[
  datamodel=acmdatamodel,
  style=acmnumeric,
  ]{biblatex}

\usepackage{cleveref}

\usepackage{times}
\usepackage{latexsym}
\usepackage{MnSymbol}
\usepackage{amsmath}
\usepackage[normalem]{ulem}
\usepackage[T1]{fontenc}
\usepackage{zi4}

\usepackage[utf8]{inputenc}

\usepackage{microtype}

\usepackage{graphicx}
\usepackage{enumitem}
\usepackage{cleveref}
\usepackage{array}
\usepackage{booktabs}
\usepackage{algorithm}
\usepackage{algpseudocode}
\usepackage{amsmath}
\usepackage{tcolorbox} 
\usepackage{multirow}
\begin{document}

\title{More Bias, Less Bias: BiasPrompting for Enhanced Multiple-Choice Question Answering}

\author{Duc Anh Vu}
\email{vuducanh001@e.ntu.edu.sg}
\affiliation{%
  \institution{Nanyang Techonological University}
  \country{Singapore}
}

\author{Nguyen Thanh Thong}
\email{e0998147@u.nus.edu}
\affiliation{%
  \institution{National University of Singapore}
  \country{Singapore}
}

\author{Cong-Duy Nguyen}
\email{duy.ntc@vinuni.edu.vn}
\affiliation{%
  \institution{Centre for AI research, VinUniversity}
  \country{Vietnam}
}

\author{Viet Anh Nguyen}
\email{nguyenvi001@e.ntu.edu.sg}
\affiliation{%
  \institution{Nanyang Techonological University}
  \country{Singapore}
}

\author{Anh Tuan Luu}
\email{anhtuan.luu@ntu.edu.sg	}
\affiliation{%
  \institution{Nanyang Techonological University}
  \country{Singapore}
}

%This command displays author info in page headers
% Please use the following convention:
% One author: J. Smith
% Two authors: J. Smith and I. Jones
% Three and more authors: J. Smith et al.
\renewcommand{\shortauthors}{Anh et al.}

\begin{abstract}

With the advancement of large language models (LLMs), their performance on multiple-choice question (MCQ) tasks has improved significantly. However, existing approaches face key limitations: answer choices are typically presented to LLMs without contextual grounding or explanation. This absence of context can lead to incomplete exploration of all possible answers, ultimately degrading the models' reasoning capabilities. To address these challenges, we introduce BiasPrompting, a novel inference framework that guides LLMs to generate and critically evaluate reasoning across all plausible answer options before reaching a final prediction. It consists of two components: first, a reasoning generation stage, where the model is prompted to produce supportive reasonings for each answer option, and then, a reasoning-guided agreement stage, where the generated reasonings are synthesized to select the most plausible answer. Through comprehensive evaluations, BiasPrompting demonstrates significant improvements in five widely used multiple-choice question answering benchmarks. Our experiments showcase that BiasPrompting enhances the reasoning capabilities of LLMs and provides a strong foundation for tackling challenging questions, particularly in settings where existing methods underperform.
\end{abstract}



% \keywords{Do, Not, Us, This, Code, Put, the, Correct, Terms, for,
% This command processes the author and affiliation and title
% information and builds the first part of the formatted document.
\maketitle

\section{Introduction}

Large language models (LLMs) have significantly advanced state-of-the-art in natural language processing (NLP) \cite{touvron2023llama, jiang2023mistral, bi2024deepseek}, excelling in tasks such as natural language understanding \cite{brown2020language, hoang-etal-2024-toxcl, wu2024fastopic, wu2024survey}, text generation \cite{zhang2024benchmarking,nguyen2022improving}, multimodal \cite{liu2023visual, nguyen-etal-2023-improving-multimodal, nguyen2025cutpaste, nguyen2024meta, nguyen2024encoding} and reasoning \cite{kojima2022large, qiao2022reasoning}. Prompt optimization has become crucial for enhancing the performance of LLMs. However, despite their capabilities, LLMs remain sensitive to input design, particularly in multiple-choice question (MCQ) settings, where they often exhibit limited instruction-following ability and shallow task understanding \cite{khatun2024study, nguyen2024video, nguyen2021enriching, nguyen2025motion}.

First, recent studies have shown that large language models are sensitive to the way input is structured, where a minor adjustment to the prompt may lead to
 an improvement or decline in performance. When provided with minimal context, such as only the answer choices without additional context, LLMs can exhibit intrinsic bias \cite{zhao2021calibrate, lum2024bias}, selection bias \cite{zheng2023large} or lack of task understanding \cite{khatun2024study}.

Second, while many prompting strategies achieve remarkable performance, they bring additional computational
burden of model inference. Prior approaches such as Chain-of-Thought (CoT) prompting \cite{wei2022chain}, Self-Refine \cite{madaan2024self} and Self-Consistency \cite{wang2022self} not only involve extensive token generation but also introduce challenges such as misinterpretation of intermediate steps and reduced interpretability \cite{yoran2023answering}. In-context learning (ICL) \cite{chen-etal-2022-improving, zhao-etal-2024-knn, vu2025curriculum} conditions the model on a long sequence of input-output demonstrations provided in the prompt. 
Furthermore, multi-agent frameworks \cite{du2023improving, liang2023encouraging, xiong2023examining} force LLMs to engage in multiple rounds of reasoning and coordination among agents before reaching a consensus.
This process significantly increases computational costs, making such methods less efficient.

\begin{figure*}[t]
    \centering
    \includegraphics[width=0.8\textwidth]{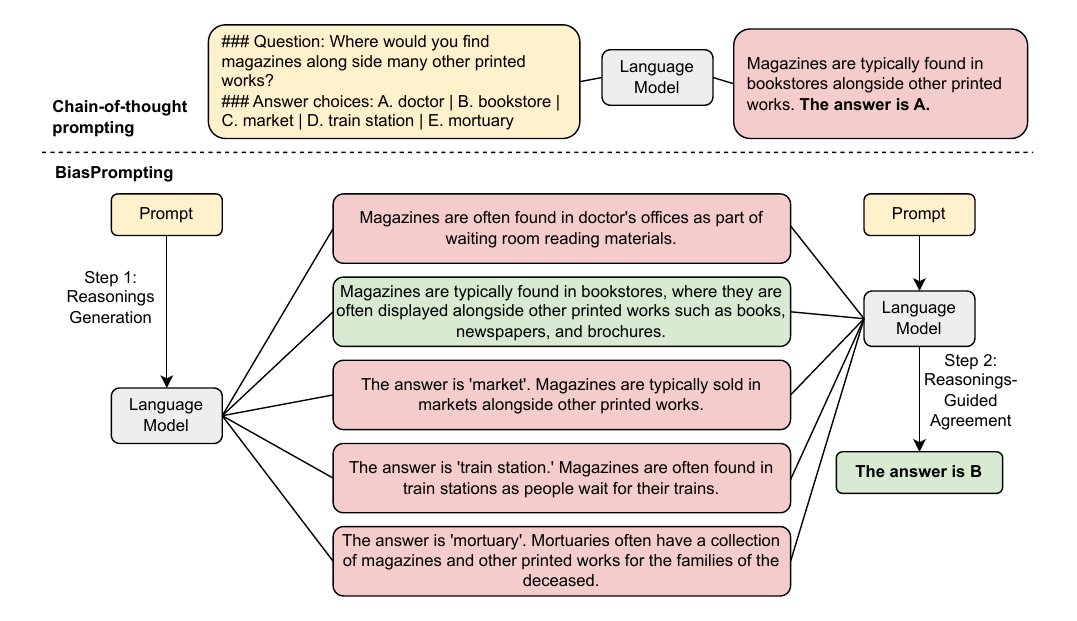}
    \caption{Illustration of BiasPrompting with two steps of Reasoning Generation and Reasonings-Guided Agreement.}
    \label{fig:biaspromptingarchit}
\end{figure*}

To address these challenges, we propose BiasPrompting, a lightweight yet effective prompting framework that enhances LLM reasoning in MCQ answering. 
% Unlike existing methods that rely on complex multi-agent frameworks or extensive token generation, BiasPrompting can operate sequentially within a single LLM, making it computationally efficient. 
BiasPrompting encourages the model to bias toward each possible candidate, enabling it to generate rich contextual reasoning for each option before achieving a final consensus. 
% It explicitly guides the LLM to engage in structured, step-by-step reasoning for each individual answer choice prior to rendering a final decision. 
This structured evaluation ensures a deeper exploration of all options, mitigating model biases and improving decision-making. In addition, unlike existing methods that rely on complex multi-agent frameworks or extensive token generation, BiasPrompting can operate sequentially within a single LLM, making it computationally efficient. Our results demonstrate significant performance gains compared to prior inference methods. Moreover, BiasPrompting uncovers latent reasoning capabilities by successfully addressing questions that other methods underperform.

\section{BiasPrompting}

% BiasPrompting is motivated by the observation that in many MCQ tasks, it is hard for LLMs to retrieve relevant facts to tackle the task given only the answer options. As shown in the example in \Cref{fig:biaspromptingarchit}, given the question \textit{"Where would you find magazines along side many other printed works?"}, LLMs have very limited additional information so as to solve the problem, only answer options with few words in this case. As a result, it outputs only related reasoning regarding its chosen option which is \textit{"doctor"}, ignoring the other underexplored answer options. 

BiasPrompting is motivated by the observation that in many multiple-choice question (MCQ) tasks, LLMs struggle to retrieve relevant facts when provided only with answer options. Formally, let an MCQ task be defined by a tuple \((Q, A, y^*)\), where \( Q \) is the question given in a natural language, \( A = \{A_1, A_2, \dots, A_n\} \) is a set of \( n \) answer options and \( y^* \in A \) is the ground-truth correct answer.
% Let \( p_\theta \) denote a pre-trained LLM with parameters \(\theta\), where \( p_\theta(y \mid Q, A) \) models the probability of selecting answer \( y \in A \) given the question and options. Traditional prompting methods directly sample \( y \sim C(p_\theta^{\text{prompt}}(y \mid Q, A)) \), often leading to biased outputs due to limited exploration of the answer choices space. 

% Let \( p_\theta \) denote a pre-trained large language model (LLM) with parameters \( \theta \), where \( p_\theta(y \mid Q, A) \textit{prompt}\) models the generated output by the LLMs. Given a question \( Q \) and candidate answers \( A \), traditional prompting methods directly predict \left[ z_{0,..,n}, y \right] = \( p_\theta(z_{0,..,n}, y \mid Q, A, \textit{prompt}) \) with y is the predicted choice and $z_{0,..,n}$ are meaningful intermediate steps toward $y$ if turned on by the frameworks.

Let \( p_\theta \) denote a pre-trained large language model (LLM) with parameters \( \theta \), where \( p_\theta(y \mid Q, A, \textit{prompt}) \) models the distribution over generated outputs conditioned on a question \( Q \), candidate answers \( A \), and a prompt design. Traditional prompting methods directly predict the sequence \( [z_0, \ldots, z_m, y] \), where \( z_0, \ldots, z_m \) are optional intermediate reasoning steps (enabled by certain prompting frameworks), and \( y \) is the final predicted answer. This sequence is modeled as:
$ \left[ z_0, \ldots, z_m, y \right] = p_\theta(z_0, \ldots, z_m, y \mid Q, A, \textit{prompt})
.$

As illustrated in \Cref{fig:biaspromptingarchit}, given the question: \textit{"Where would you find magazines along side many other printed works?"}, LLMs are given a limited set of answer options, each consisting of only a few words. Due to this constraint, the model often generates reasoning for its chosen option only, ignoring underexplored alternatives. In this example, the model selects \textit{"doctor"} as the answer, failing to consider other options meaningfully.

% In this case, BiasPrompting first forces the LLM to generate reasonings for each individual answer option (\textit{"doctor", "bookstore", "market", "train station", "mortuary"}), arguing in favor of each option as if defending it, regardless of its correctness. This step ensures that the model retrieves all potentially relevant information before making a final decision. By systematically introducing controlled biases in the prompt design, we hypothesize that this approach mitigates biases in LLMs towards favoring specific answers. BiasPrompting consists of two fundamental steps:

BiasPrompting first prompts the LLM to generate supporting reasoning for each answer option (e.g., \textit{"doctor", "bookstore", "market", "train station", "mortuary"}), \textbf{treating each as if it were correct}. This encourages the model to retrieve a broad range of relevant information before selecting an answer. By embedding controlled biases into the prompt, we hypothesize this method reduces the LLM’s tendency to favor specific options. BiasPrompting involves two key steps:

% \paragraph{Reasoning Generation}

% Given a multiple-choice question \( Q \) with a set of answer choices \( A = \{A_1, A_2, \dots, A_n\} \), the model first generates a corresponding reasoning \( R_i \) for each \( A_i \) separately. Unlike conventional prompting methods that seek to determine the correctness of an answer, BiasPrompting explicitly instructs the model to generate reasoning that supports each answer choice, regardless of its validity. This structured approach prevents the model from prematurely discarding answer choices and encourages a more exhaustive exploration of possible responses. Even if incorrect explanations are generated, this process fosters a more holistic understanding of all answer choices, ultimately improving the quality of the final prediction. An example prompt for reasoning generation is shown in \Cref{tab:reasoning_gen_prompt}.
\subsection{Reasoning Generation}

Given an MCQ task with question \( Q \) and answer options \( A \), BiasPrompting first prompts the LLM to generate a reasoning \( R_i \) for each answer option \( A_i \), where \( i = 1, 2, \dots, n \). Unlike prior methods that directly evaluate the correctness of each option, BiasPrompting instructs the model to produce reasoning that supports \( A_i \) as if it were the correct answer, regardless of its actual validity. This process is formalized as:
% \(\left R_i \sim p_\theta(R_i \mid Q, A_i, \textit{prompt}_{\text{RG}}) \right\), where \( \textit{prompt}_{\text{RG}} \)  is a prompt designed to generate reasoning that supports $\left A_i \right$. 
\( R_i \sim p_\theta(R_i \mid Q, A_i, \textit{prompt}_{\text{RG}}) \),
where \( \textit{prompt}_{\text{RG}} \) is a prompt designed to generate reasoning that supports \( A_i \).

% For example, for the option \textit{"doctor"}, the prompt might ask the model to explain why a doctor's office could plausibly contain magazines alongside other printed works.
This structured approach ensures that the model explores all answer options exhaustively, preventing premature dismissal of alternatives. Even if some generated reasonings are incorrect, this step enriches the model’s understanding of the context and improves the robustness of the final prediction. The template of \( \textit{prompt}_{\text{RG}} \) is provided in \Cref{tab:reasoning_gen_prompt}.

% \paragraph{Reasonings-Guided Agreement}

% Once reasoning has been generated for all answer choices, the model constructs a structured prompt that includes the original question \( Q \), the complete set of answer choices \( A \), and the generated reasonings \( R = \{R_1, R_2, \dots, R_n\} \). This prompt is then used to predict the final answer \( P \). By incorporating explicitly generated reasoning into the decision-making process, the model gains additional context beyond simply selecting from the given answer choices. 
% % This step enhances the interpretability of the model’s decision, making its predictions more transparent and well-grounded in reasoning.

% An example prompt used in BiasPrompting is shown in \Cref{tab:biaspromptprompt}. Following \citet{long2024llms}, we use a placeholder format to wrap the final selected answer in multiple-choice questions. Additionally, we explore prompting techniques such as zero-shot and chain-of-thought \cite{wei2022chain} to enhance model's performance in this stage. Further details on this are discussed in \Cref{res}.

\begin{table*}[t]
\centering
\renewcommand{\arraystretch}{1.1}
\resizebox{\linewidth}{!}{
\begin{tabular}{c|l|ccccc}
\toprule
 & \textbf{Mode} & \textbf{CommonsenseQA} & \textbf{StrategyQA} & \textbf{PIQA} & \textbf{Date Understanding} & \textbf{Causal Judgement} \\
\midrule
\multirow{3}{*}{\rotatebox{90}{\textbf{Mistral}}}
 & Zero-shot & 65.1 & 63.8 & \textbf{75.1} & 35.6 & 53.5 \\
 & Chain-of-Thought & \textbf{67.1} \textcolor{green!60!black}{\scriptsize(+2.0\%)} & 52.8 \textcolor{red}{\scriptsize(-11.0\%)} & 70.5 \textcolor{red}{\scriptsize(-4.6\%)} & 32.0 \textcolor{red}{\scriptsize(-3.6\%)} & 54.6 \textcolor{green!60!black}{\scriptsize(+1.1\%)} \\
 & BiasPrompting & 66.0 \textcolor{green!60!black}{\scriptsize(+0.9\%)} & \textbf{67.7} \textcolor{green!60!black}{\scriptsize(+3.9\%)} & 74.2 \textcolor{red}{\scriptsize(-0.9\%)} & \textbf{40.0} \textcolor{green!60!black}{\scriptsize(+4.4\%)} & \textbf{62.6} \textcolor{green!60!black}{\scriptsize(+9.1\%)} \\
\midrule
\multirow{3}{*}{\rotatebox{90}{\textbf{Deepseek}}}
 & Zero-shot & \textbf{50.1} & 57.1 & 64.8 & 37.6 & 53.5 \\
 & Chain-of-Thought & 49.3 \textcolor{red}{\scriptsize(-0.8\%)} & 52.7 \textcolor{red}{\scriptsize(-4.4\%)} & 63.8 \textcolor{red}{\scriptsize(-1.1\%)} & 36.4 \textcolor{red}{\scriptsize(-1.2\%)} & 49.7 \textcolor{red}{\scriptsize(-3.8\%)} \\
 & BiasPrompting & 48.5 \textcolor{red}{\scriptsize(-1.6\%)} & \textbf{57.5} \textcolor{green!60!black}{\scriptsize(+0.4\%)} & \textbf{64.9} \textcolor{green!60!black}{\scriptsize(+0.1\%)} & \textbf{40.0} \textcolor{green!60!black}{\scriptsize(+2.4\%)} & \textbf{55.6} \textcolor{green!60!black}{\scriptsize(+2.1\%)} \\
\midrule
\multirow{3}{*}{\rotatebox{90}{\textbf{Gemma}}}
 & Zero-shot & 27.0 & 48.5 & 48.9 & 20.0 & \textbf{51.9} \\
 & Chain-of-Thought & 38.0 \textcolor{green!60!black}{\scriptsize(+11.0\%)} & 44.4 \textcolor{red}{\scriptsize(-4.1\%)} & 51.0 \textcolor{green!60!black}{\scriptsize(+2.1\%)} & \textbf{27.6} \textcolor{green!60!black}{\scriptsize(+7.6\%)} & 38.0 \textcolor{red}{\scriptsize(-13.9\%)} \\
 & BiasPrompting & \textbf{38.7} \textcolor{green!60!black}{\scriptsize(+11.7\%)} & \textbf{49.1} \textcolor{green!60!black}{\scriptsize(+0.6\%)} & \textbf{54.2} \textcolor{green!60!black}{\scriptsize(+5.3\%)} & 23.2 \textcolor{green!60!black}{\scriptsize(+3.2\%)} & \textbf{51.9} \textcolor{red}{\scriptsize(-0.0\%)} \\
\bottomrule
\end{tabular}
}
\caption{Performance comparison across datasets and models with greedy decoding. Note that there can
 be an error of 0.1 due to rounding.}
\label{tab:results}
\end{table*}

\subsection{Reasonings-Guided Agreement}

After generating reasonings \( R = \{R_1, R_2, \dots, R_n\} \) for all answer options, BiasPrompting constructs a final prompt that integrates the original question \( Q \), the answer options \( A \), and the generated reasonings \( R \). The LLM then predicts the final answer \( y \in A \), modeled as:
\( y \sim p_\theta(y \mid Q, A, R, \textit{prompt}_{\text{consensus}}) \) where \( \textit{prompt}_{\text{consensus}} \) is designed to synthesize the reasonings and select the most appropriate answer. This prompt provides the model with richer context, enabling a more informed and transparent decision-making process compared to direct answer selection. The template of \( \textit{prompt}_{\text{consensus}} \) is shown in \Cref{tab:biaspromptprompt}. Following \cite{long2024llms}, we adopt a placeholder format to wrap the selected answer in MCQ tasks. Additionally, we explore prompting techniques such as zero-shot and chain-of-thought \cite{wei2022chain} to enhance performance in this stage. Further details are discussed in \Cref{res}.

\section{Experiment}

\subsection{Datasets} We cover abroad set of diverse MCQ tasks, varying length and domain. Our evaluation covers five popular multiple-choice question benchmarks: (i) CommonsenseQA \cite{talmor-etal-2019-commonsenseqa}, (ii) StrategyQA \cite{geva2021did}, (iii) PIQA \cite{bisk2020piqa}, (iv) BBH–Date Understanding, and (v) BBH–Causal Judgement \cite{suzgun2022challenging}. These datasets primarily test commonsense reasoning and vary in the number of answer options per question, ranging from 2 to 6. Detailed dataset statistics are provided in \Cref{data_stat}.

\begin{table}[h]
\centering
\small
\renewcommand{\arraystretch}{1}
\begin{tabular}{lcc}
\toprule
\textbf{Dataset}    & \textbf{Task Type}              & \textbf{Size} \\ \midrule

CSQA \cite{talmor-etal-2019-commonsenseqa} & 5 Choices & 1.221 \\
StrategyQA \cite{geva2021did}& 2 Choices & 2,290 \\
 PIQA \cite{bisk2020piqa}& 2 Choices & 1,838 \\
BBH-DU \cite{suzgun2022challenging} & 6 Choices & 250 \\
BBH--CJ \cite{suzgun2022challenging} & 2 Choices & 187 \\ \bottomrule
\end{tabular}
\caption{Testing dataset statistics.}
\label{data_stat}
\end{table}

% \subsection{Setup} We evaluate our method using three widely used, high-performing large language models (LLMs): Mistral-7B \cite{jiang2023mistral}, DeepSeek-7B \cite{bi2024deepseek}, and Gemma-7B \cite{Mesnard2024GemmaOM}.

\begin{figure*}[t]

    \centering
    \includegraphics[width=0.9\textwidth]{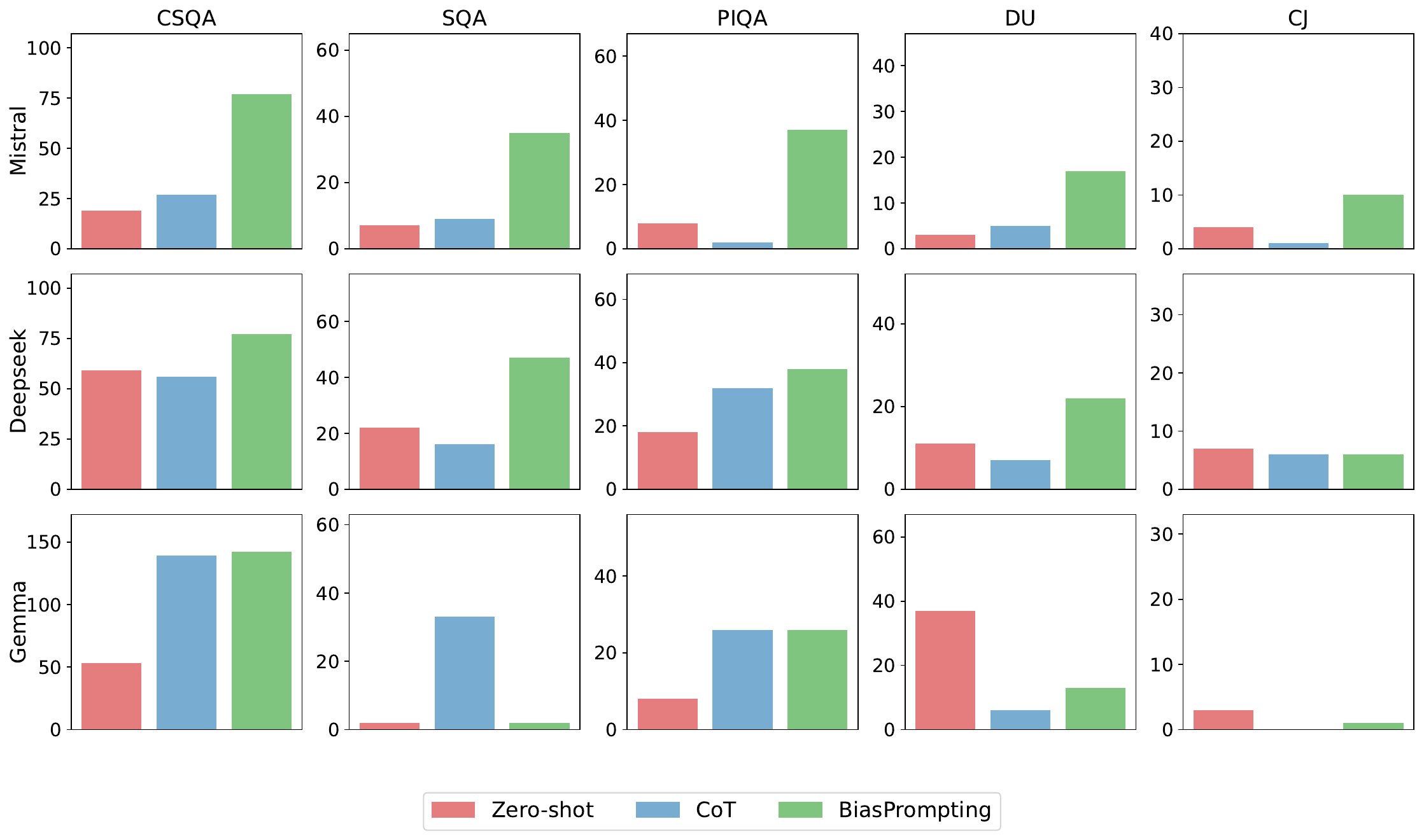}
    \caption{Number of questions successfully answered by each prompting method while the other two underperform.}

    \label{fig:vv}
\end{figure*}

\subsection{Model Details and Hyperparameters}

We evaluate our method using three widely used, high-performing large language models (LLMs):

\begin{itemize}
    \item Mistral-7B-It-v0.2\footnote{\url{https://huggingface.co/mistralai/Mistral-7B-Instruct-v0.2}}\cite{jiang2023mistral}: A 7B-parameter decoder-only model with an 8k context window, outperforming LLaMA-2-13B \cite{touvron2023llama} on multiple benchmarks.

\item DeepSeek-7B-chat\footnote{\url{https://huggingface.co/deepseek-ai/deepseek-llm-7b-chat}}\cite{bi2024deepseek}: A 7B-parameter decoder-only model trained from scratch on a large-scale corpus of 2 trillion tokens.

\item Gemma-7B-It\footnote{\url{https://huggingface.co/google/gemma-7b-it}}\cite{Mesnard2024GemmaOM}: A 7B-parameter decoder-only model with an 8k context window, noted for strong reasoning and problem-solving capabilities.

\end{itemize}
We investigate the open-source models through
HuggingFace transformers library \cite{wolf-etal-2020-transformers}. To avoid empty or truncated outputs, we set a non-zero minimum number of generated tokens. For all models, we fix the context window at 1024 tokens and employ greedy decoding as the generation strategy.

\subsection{Evaluation Measures}

We report accuracy for all datasets. During inference, models are instructed to produce answers in a specific format to enable automated parsing of the predicted choice. A prediction is counted as correct if the parsed answer exactly matches the ground-truth label. We assess statistical significance of accuracy differences using a two-proportion $z$-test for each dataset–model pair, considering results significant when $p<0.05$. All reported improvements meet this criterion.

\subsection{Results \& Discussions}
\label{res}
\subsubsection{Main results}

\Cref{tab:results} provides a comprehensive comparison of
 various models on five MCQ datasets. Overall, BiasPrompting consistently achieves superior performance, outperforming the zero-shot baseline in most cases. While zero-shot CoT sometimes attains the highest accuracy across certain datasets, its performance is highly inconsistent, often experiencing significant drops. In contrast, BiasPrompting offers a more stable and reliable improvement in all datasets, where it achieves the best or near-best scores. Moreover, unlike CoT, which depends on detailed step-by-step reasoning that can result in lengthy responses, BiasPrompting offers a more lightweight approach, improving performance without the need for extensive reasoning chains. Given these findings, BiasPrompting emerges as a promising strategy for mitigating inconsistencies observed in zero-shot CoT, offering both stability and performance gains while maintaining efficiency.

\subsubsection{Latent reasoning capabilities with BiasPrompting}

To further qualitatively demonstrate the superiority of BiasPrompting, we evaluate the results on the questions where baseline methods underperform. As illustrated in \Cref{fig:vv}, the number of questions that the other two approaches fail to address across the majority of experimental cases under BiasPrompting is significantly higher than Zero-shot and CoT. This highlights the latent reasoning capabilities of incorporating reasoning for all answer choices within the prompt, enabling the model to better evaluate and differentiate between options. Furthermore, BiasPrompting reveals previously unexplored capabilities of a single LLM, demonstrating its potential to enhance reasoning without requiring multiple LLMs or extensive prompting techniques.

\begin{table}[H]
\centering
\renewcommand{\arraystretch}{1}
\begin{tabular}{@{}l|l|ccc@{}}
\toprule
\textbf{Model}    & \textbf{Mode}              & \textbf{CSQA} & \textbf{SQA}  & \textbf{PIQA} \\ \midrule
\textbf{Mistral}           & +Bias            & 66.01         & \textbf{67.69} & \textbf{74.16 }        \\
                  & +Bias +CoT      &\textbf{ 66.50 }        & 66.37         & 73.55         \\ \midrule
\textbf{Deepseek}          & +Bias            & 46.68         & \textbf{57.50} & 64.91 \\
                  & +Bias +CoT      &\textbf{ 46.92 }        & 46.14         & \textbf{65.77  }       \\ \bottomrule
\end{tabular}
\caption{Performance comparison of Mistral and Deepseek models under BiasPrompting with and without CoT in the predicting final answer stage. Bold indicates the best scores among all methods.}
\label{tab:biascot}
\end{table}

\subsubsection{BiasPrompting + CoT}

We investigate the impact of incorporating Chain-of-Thought (CoT) reasoning into the Reasonings-Guided Agreement framework. In this setup, the language model generates intermediate reasoning steps in addition to the final answer $\left( y \in A \right)$, modeled as: 
\[
\left[ z_0, \ldots, z_m, y \right] = p_\theta(z_0, \ldots, z_m, y \mid Q, A, R, \textit{prompt}_{\text{consensus}})
\]

% As shown in \Cref{tab:biascot}, incorporating CoT alongside BiasPrompting does not yield a significant improvement in the final answer prediction stage. While there are minor variations in performance, the results indicate that adding CoT does not consistently enhance accuracy across different datasets. In some cases, such as in the Mistral model for CSQA, the combination of BiasPrompting and CoT slightly underperforms compared to BiasPrompting alone. These findings imply that BiasPrompting alone is sufficient for improving model performance, and the additional complexity introduced by CoT does not necessarily translate into better results. 
As shown in \Cref{tab:biascot}, integrating CoT with BiasPrompting does not lead to significant improvements in final answer prediction. While minor performance fluctuations are observed, the overall results suggest that CoT does not consistently enhance accuracy across datasets. In some cases, such as the Mistral model on the CSQA, the combination of BiasPrompting and CoT even underperforms compared to BiasPrompting alone. These findings demonstrate that BiasPrompting alone is sufficient for improving performance, and the added complexity from CoT does not necessarily translate to better outcomes.

\begin{figure*}[]
    
    \centering
    \includegraphics[width=\textwidth]{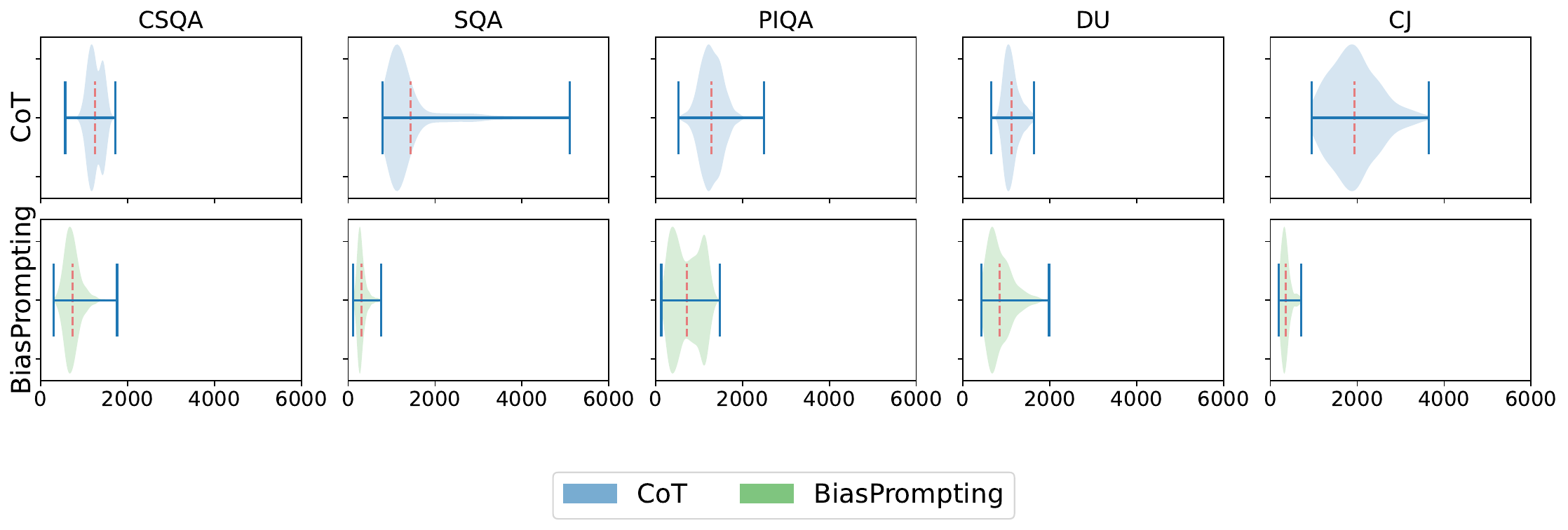}
    \caption{Violin plots comparing the average generated tokens from BiasPrompting and CoT prompting across five datasets (CSQA, SQA, PIQA, DU, and CJ) using the Gemma model. The {\color[HTML]{86b66c} $\blacksquare$} violins represent BiasPrompting, while the {\color[HTML]{78acd1} $\blacksquare$} violins correspond to CoT prompting. The dashed line inside each violin indicates the mean generated tokens.}
    \label{fig:avglength}
\end{figure*}

\subsubsection{Token Efficiency Analysis}

To assess computational efficiency, we compare the total number of generated tokens for BiasPrompting and CoT across five datasets, aggregating tokens over all rounds of prompting required to produce a final answer (\Cref{fig:avglength}). The results show that BiasPrompting consistently generates substantially fewer tokens than CoT, while maintaining or improving accuracy. This reduction stems from its single-pass design, which integrates reasoning for all answer choices within one prompt, avoiding the lengthy, multi-round reasoning chains typical of CoT. Consequently, BiasPrompting achieves both inference-time efficiency and performance gains, highlighting its practicality for deployment in resource-constrained settings.

\subsubsection{BiasPrompting Reduces Selection Bias}
\begin{figure}[H]

    \centering
    \includegraphics[width=0.37\textwidth]{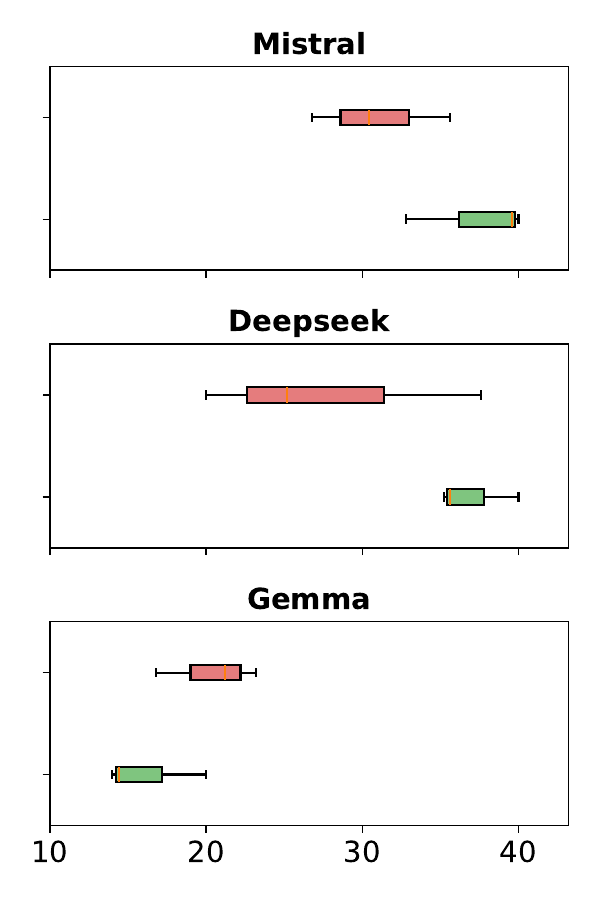}
    \caption{Results of option order swapping on zero-shot and BiasPrompting performance across random option orderings. {\color[HTML]{86b66c} $\blacksquare$} denotes the box-plot of BiasPrompting, while {\color[HTML]{c7716e} $\blacksquare$}  represents the box-plot of  zero-shot. }
    \label{fig:swaporder}
\end{figure}

\Cref{fig:swaporder} presents the robustness comparison between Zero-shot and BiasPrompting across three random answer option orderings for Mistral, DeepSeek, and Gemma. In all cases, BiasPrompting yields higher median accuracy and lower variance than Zero-shot, demonstrating improved overall performance. These results suggest that BiasPrompting effectively reduces the susceptibility of model predictions to option order permutations, addressing the selection bias highlighted by \cite{zheng2023large}.

% \subsection{}

% \subsection{MCQs solving}

\section{Related Work}

\subsection{In-context learning}

Numerous studies have focused on enhancing LLM performance through improved prompting techniques. Early research such as Chain-of-Thought (CoT) \cite{wei2022chain}, Thread-of-Thought \cite{zhou2023thread}, and Tab-CoT \cite{jin2023tab} encourage LLMs to explicitly generate intermediate reasoning steps before producing a final answer, improving their ability to solve complex problems. Additionally, ensemble-based approaches \cite{wang2022self, madaan2024self, khalifa2023exploring} leverage self-consistency by generating multiple responses and selecting the most frequent one, effectively reducing output variance and improving reliability. Another line of research has explored problem decomposition, where complex tasks are broken down into simpler sub-questions to enhance reasoning accuracy \cite{khot2022decomposed, lyu2023faithful, zhou2022least}. 

Beyond these structured prompting techniques, various zero-shot prompting strategies have been developed to guide LLMs more effectively. Persona prompting \cite{wang2023rolellm} assigns a specific role to the model, influencing its response style, while emotion-based prompting \cite{li2023large} incorporates psychologically relevant phrases to elicit more human-like and context-aware outputs. Recent extensions, such as  \cite{zhao2021calibrate}, adjust for inherent biases in zero-shot predictions, further stabilizing performance. Few-shot prompting builds on zero-shot by incorporating a small number of examples in the prompt, enabling in-context learning (ICL) without fine-tuning, which conditions the model on input-output demonstrations, allowing adaptation to new tasks with minimal data. For example, \cite{chen2022improvingincontextfewshotlearning} demonstrates how self-supervised training enhances few-shot ICL, improving generalization in low-data scenarios. Recent advancements refine ICL for cost and effectiveness. Techniques like Decomposed Prompting \cite{khot2023decomposedpromptingmodularapproach} break examples into sub-tasks, while Demonstration Ensembling \cite{khalifa2023exploringdemonstrationensemblingincontext} combines multiple few-shot samples for robustness. 

ICL variants also address domain-specific challenges. For instance, \cite{vu2024curriculumdemonstrationselectionincontext} introduces curriculum demonstration selection, progressively increasing example complexity to boost few-shot performance. In educational applications, \cite{medical} uses few-shot prompts for generating language-agnostic MCQs, leveraging chain-of-thought-inspired multi-stage prompting to create diverse distractors. Despite its strengths, few-shot ICL can amplify biases from examples \cite{lum2025biaslanguagemodelstrick} and struggles with long contexts due to token limits. Mitigation efforts have been made to these issues, such as \cite{zhao-etal-2024-knn} explores k-nearest neighbor ICL for compositional generalization.

% \subsection{Bias and fairness in LLMs} 

% Inherent bias in LLMs can perpetuate harmful stereotypes, particularly in MCQ tasks. Intrinsic biases arise from imbalanced datasets \cite{ranjan2024comprehensivesurveybiasllms}), while extrinsic biases manifest in outputs \cite{long2024llms}. Surveys like \cite{gallegos2024biasfairnesslargelanguage} categorize harms into representational (e.g., stereotypes) and allocational (e.g., unfair decisions), proposing metrics at embedding, probability, and generation levels. 

% Mitigation directions for bias include pre-processing (e.g., data augmentation \cite{lyu2023faithful}), in-training (e.g., fairness constraints\cite{ranjan2024comprehensivesurveybiasllms}), and post-processing (e.g., self-diagnosis \cite{madaan2024self}). For MCQ answering, single-token logit prompting \cite{cappelletti2025improvingllmfirsttokenpredictions} integrates retrieval-augmented generation to counter shuffled options. 

\subsection{LLMs on MCQ answering}

Prior works have explored MCQ solving from multiple research angles. For example, in terms of LLM inherent bias, \cite{zheng2023large} identified and analyzed selection bias, showing that changes in the ordering of answer options can significantly alter model predictions, similarly,\cite{pezeshkpour-hruschka-2024-large} studied positional bias, where certain positions in the answer list are favored regardless of content.

Beyond bias analysis, prompt engineering approaches such as \cite{long2024llms, liu2024we} propose optimally designed templates tailored for MCQ settings to improve consistency and accuracy. Prior works also conduct in-depth investigations into the specific behaviors exhibited by LLMs when the underlying task is MCQ solving, for instance, \cite{wang2024llmsperformmcqaselecting} reveals that LLMs often succeed in MCQA not through true understanding but by eliminating implausible options, highlighting a gap between apparent performance and actual reasoning depth. In specialized domains, \cite{tomova2024leveraging} uses LLMs for medical MCQ feedback, while \cite{wang2024llmsperformmcqaselecting} critiques MCQ rationality, advocating true-or-false complements.

\section{Conclusion}

% In this paper, we introduced BiasPrompting, a novel prompting strategy that enhances LLM performance in multiple-choice question answering by systematically generating reasoning for all answer choices before making a final selection. Through extensive experiments across diverse datasets and models, we demonstrated that BiasPrompting consistently improves accuracy, enhances stability, and mitigates model bias compared to traditional approaches. Our findings highlight its ability to improve decision-making in challenging questions where existing methods often struggle.

In this paper, we introduced BiasPrompting, a novel inference framework that enhances LLM performance in multiple-choice question answering by generating reasonings for all answer choices before selecting a final answer. Our experiments demonstrate that BiasPrompting consistently improves accuracy, mitigates biases like ordering effects, and offers greater stability compared to traditional methods while being more computationally efficient. Our findings also highlight its ability to improve decision-making in challenging questions where other methods often underperform. We hope our work will encourage more human-like approaches to unlock the hidden potential of LLMs.

\section{Limitations}

While BiasPrompting demonstrates significant improvements in multiple-choice question answering, several limitations remain that warrant further investigation. First, our experiments primarily focus on commonsense reasoning tasks. While these benchmarks are valuable for assessing LLM reasoning capabilities, BiasPrompting has not been systematically evaluated on other reasoning domains. Extending BiasPrompting to a broader range of tasks would provide deeper insights into its generalizability and effectiveness across different reasoning paradigms. Second, our study is limited to 7B parameter models (Mistral-7B, DeepSeek-7B, and Gemma-7B). While these models offer a balance between computational efficiency and performance, they may not fully capture the behavior of larger-scale LLMs. The impact of BiasPrompting on these larger models remains unexplored and could reveal new insights into its scalability, efficiency, and effectiveness at different model sizes. Additionally, the quality of reasoning generation has not been extensively refined. Future work can focus on enhancing reasoning generation to improve accuracy and depth without compromising efficiency.

\appendix

\section{Prompts used in BiasPrompting}
\begin{table}[H]
\centering
\footnotesize
\tcbset{
    tabularbox/.style={colback=white, colframe=black, boxrule=0.1pt, arc=5mm}
}
\begin{tcolorbox}[tabularbox]
\begin{tabular}{p{0.9\textwidth}}
\raggedright
Given the following question:\newline A revolving door is convenient for two direction travel, but it also serves as a security measure at a what?\newline Provide reasoning proving that 'bank' is the correct choice without any textual description in one sentence.
\end{tabular}
\end{tcolorbox}
\caption{Reasoning generation prompt example}
\label{tab:reasoning_gen_prompt}
\end{table}

% \section{Prompts used for BiasPrompting}

\begin{table}[H]
\centering
\footnotesize
\tcbset{
    tabularbox/.style={colback=white, colframe=black, boxrule=0.1pt, arc=5mm}
}
\begin{tcolorbox}[tabularbox]
\begin{tabular}{p{0.9\textwidth}}
\raggedright

\#\#\# Question: A revolving door is convenient for two direction travel, but it also serves as a security measure at a what? \newline
\#\#\# Answer choices: A. bank | B. library | C. department store | D. mall | E. new york \newline
\#\#\# Reasoning for answer choice A: The answer is 'bank'. Revolving doors are often used in banks to control the flow of people entering and exiting. \newline
\#\#\# Reasoning for answer choice B: The answer is 'library'. Revolving doors are commonly used in libraries to control the flow of people entering and exiting.\newline
\#\#\# Reasoning for answer choice C: The answer is department store. Revolving doors are commonly used in department stores to control the flow of customers.\newline
\#\#\# Reasoning for answer choice D: The answer is 'mall'.
Revolving doors are commonly found in malls to control pedestrian traffic flow and prevent congestion.\newline
\#\#\# Reasoning for answer choice E: The answer is New York. The revolving door is a security measure at the Empire State Building in New York City.\newline

Wrap your final answer by filling in the placeholder below: 'So the answer is: \{\{placeholder\}\}'
\end{tabular}
\end{tcolorbox}
\caption{BiasPrompting prompt example}
\label{tab:biaspromptprompt}
\end{table}

\printbibliography

@String{Computing = "Computing" }

@String{Springer = "Springer-Verlag" }

@ArtifactSoftware{R,
    title = {R: A Language and Environment for Statistical Computing},
    author = {{R Core Team}},
    organization = {R Foundation for Statistical Computing},
    address = {Vienna, Austria},
    year = {2019},
    url = {https://www.R-project.org/},
}

@article{touvron2023llama,
  title={Llama: Open and efficient foundation language models},
  author={Touvron, Hugo and Lavril, Thibaut and Izacard, Gautier and Martinet, Xavier and Lachaux, Marie-Anne and Lacroix, Timoth{\'e}e and Rozi{\`e}re, Baptiste and Goyal, Naman and Hambro, Eric and Azhar, Faisal and others},
  journal={arXiv preprint arXiv:2302.13971},
  year={2023}
}

@article{long2024llms,
  title={Llms are biased towards output formats! systematically evaluating and mitigating output format bias of llms},
  author={Long, Do Xuan and Ngoc, Hai Nguyen and Sim, Tiviatis and Dao, Hieu and Joty, Shafiq and Kawaguchi, Kenji and Chen, Nancy F and Kan, Min-Yen},
  journal={arXiv preprint arXiv:2408.08656},
  year={2024}
}

@article{wei2022chain,
  title={Chain-of-thought prompting elicits reasoning in large language models},
  author={Wei, Jason and Wang, Xuezhi and Schuurmans, Dale and Bosma, Maarten and Xia, Fei and Chi, Ed and Le, Quoc V and Zhou, Denny and others},
  journal={Advances in neural information processing systems},
  volume={35},
  pages={24824--24837},
  year={2022}
}

@article{bi2024deepseek,
  title={Deepseek llm: Scaling open-source language models with longtermism},
  author={Bi, Xiao and Chen, Deli and Chen, Guanting and Chen, Shanhuang and Dai, Damai and Deng, Chengqi and Ding, Honghui and Dong, Kai and Du, Qiushi and Fu, Zhe and others},
  journal={arXiv preprint arXiv:2401.02954},
  year={2024}
}

@article{Mesnard2024GemmaOM,
  title={Gemma: Open Models Based on Gemini Research and Technology},
  author={Gemma Team Thomas Mesnard and Cassidy Hardin and Robert Dadashi and Surya Bhupatiraju and Shreya Pathak and L. Sifre and Morgane Rivi{\`e}re and Mihir Kale and J Christopher Love and Pouya Dehghani Tafti and L'eonard Hussenot and Aakanksha Chowdhery and Adam Roberts and Aditya Barua and Alex Botev and Alex Castro-Ros and Ambrose Slone and Am'elie H'eliou and Andrea Tacchetti and Anna Bulanova and Antonia Paterson and Beth Tsai and Bobak Shahriari and Charline Le Lan and Christopher A. Choquette-Choo and Cl{\'e}-ment Crepy and Daniel Cer and Daphne Ippolito and David Reid and Elena Buchatskaya and Eric Ni and Eric Noland and Geng Yan and George Tucker and George-Christian Muraru and Grigory Rozhdestvenskiy and Henryk Michalewski and Ian Tenney and Ivan Grishchenko and Jacob Austin and James Keeling and Jane Labanowski and Jean-Baptiste Lespiau and Jeff Stanway and Jenny Brennan and Jeremy Chen and Johan Ferret and Justin Chiu and Justin Mao-Jones and Kather-ine Lee and Kathy Yu and Katie Millican and Lars Lowe Sjoesund and Lisa Lee and Lucas Dixon and Machel Reid and Maciej Mikuła and Mateo Wirth and Michael Sharman and Nikolai Chinaev and Nithum Thain and Olivier Bachem and Oscar Chang and Oscar Wahltinez and Paige Bailey and Paul Michel and Petko Yotov and Pier Giuseppe Sessa and Rahma Chaabouni and Ramona Comanescu and Reena Jana and Rohan Anil and Ross McIlroy and Ruibo Liu and Ryan Mullins and Samuel L. Smith and Sebastian Borgeaud and Sertan Girgin and Sholto Douglas and Shree Pandya and Siamak Shakeri and Soham De and Ted Klimenko and Tom Hennigan and Vladimir Feinberg and Wojciech Stokowiec and Yu-hui Chen and Zafarali Ahmed and Zhitao Gong and Tris Warkentin and Ludovic Peran and Minh Giang and Cl{\'e}ment Farabet and Oriol Vinyals and Jeffrey Dean and Koray Kavukcuoglu and Demis Hassabis and Zoubin Ghahramani and Douglas Eck and Joelle Barral and Fernando Pereira and Eli Collins and Armand Joulin and Noah Fiedel and Evan Senter and Alek Andreev and Kathleen Kenealy},
  journal={ArXiv},
  year={2024},
  volume={abs/2403.08295},
  url={https://api.semanticscholar.org/CorpusID:268379206}
}

@inproceedings{talmor-etal-2019-commonsenseqa,
    title = "{C}ommonsense{QA}: A Question Answering Challenge Targeting Commonsense Knowledge",
    author = "Talmor, Alon  and
      Herzig, Jonathan  and
      Lourie, Nicholas  and
      Berant, Jonathan",
    editor = "Burstein, Jill  and
      Doran, Christy  and
      Solorio, Thamar",
    booktitle = "Proceedings of the 2019 Conference of the North {A}merican Chapter of the Association for Computational Linguistics: Human Language Technologies, Volume 1 (Long and Short Papers)",
    month = jun,
    year = "2019",
    address = "Minneapolis, Minnesota",
    publisher = "Association for Computational Linguistics",
    url = "https://aclanthology.org/N19-1421/",
    doi = "10.18653/v1/N19-1421",
    pages = "4149--4158"
}

@article{geva2021did,
  title={Did aristotle use a laptop? a question answering benchmark with implicit reasoning strategies},
  author={Geva, Mor and Khashabi, Daniel and Segal, Elad and Khot, Tushar and Roth, Dan and Berant, Jonathan},
  journal={Transactions of the Association for Computational Linguistics},
  volume={9},
  pages={346--361},
  year={2021},
  publisher={MIT Press One Rogers Street, Cambridge, MA 02142-1209, USA journals-info~…}
}

@inproceedings{bisk2020piqa,
  title={Piqa: Reasoning about physical commonsense in natural language},
  author={Bisk, Yonatan and Zellers, Rowan and Gao, Jianfeng and Choi, Yejin and others},
  booktitle={Proceedings of the AAAI conference on artificial intelligence},
  volume={34},
  number={05},
  pages={7432--7439},
  year={2020}
}

@article{suzgun2022challenging,
  title={Challenging big-bench tasks and whether chain-of-thought can solve them},
  author={Suzgun, Mirac and Scales, Nathan and Sch{\"a}rli, Nathanael and Gehrmann, Sebastian and Tay, Yi and Chung, Hyung Won and Chowdhery, Aakanksha and Le, Quoc V and Chi, Ed H and Zhou, Denny and others},
  journal={arXiv preprint arXiv:2210.09261},
  year={2022}
}

@article{jiang2023mistral,
  title={Mistral 7B},
  author={Jiang, Albert Q and Sablayrolles, Alexandre and Mensch, Arthur and Bamford, Chris and Chaplot, Devendra Singh and Casas, Diego de las and Bressand, Florian and Lengyel, Gianna and Lample, Guillaume and Saulnier, Lucile and others},
  journal={arXiv preprint arXiv:2310.06825},
  year={2023}
}

@article{zheng2023large,
  title={On Large Language Models' Selection Bias in Multi-Choice Questions},
  author={Zheng, Chujie and Zhou, Hao and Meng, Fandong and Zhou, Jie and Huang, Minlie},
  journal={arXiv preprint arXiv:2309.03882},
  year={2023}
}

@article{khatun2024study,
  title={A Study on Large Language Models' Limitations in Multiple-Choice Question Answering},
  author={Khatun, Aisha and Brown, Daniel G},
  journal={arXiv preprint arXiv:2401.07955},
  year={2024}
}

@article{zhang2024benchmarking,
  title={Benchmarking large language models for news summarization},
  author={Zhang, Tianyi and Ladhak, Faisal and Durmus, Esin and Liang, Percy and McKeown, Kathleen and Hashimoto, Tatsunori B},
  journal={Transactions of the Association for Computational Linguistics},
  volume={12},
  pages={39--57},
  year={2024},
  publisher={MIT Press One Broadway, 12th Floor, Cambridge, Massachusetts 02142, USA~…}
}

@article{kojima2022large,
  title={Large language models are zero-shot reasoners},
  author={Kojima, Takeshi and Gu, Shixiang Shane and Reid, Machel and Matsuo, Yutaka and Iwasawa, Yusuke},
  journal={Advances in neural information processing systems},
  volume={35},
  pages={22199--22213},
  year={2022}
}

@article{qiao2022reasoning,
  title={Reasoning with language model prompting: A survey},
  author={Qiao, Shuofei and Ou, Yixin and Zhang, Ningyu and Chen, Xiang and Yao, Yunzhi and Deng, Shumin and Tan, Chuanqi and Huang, Fei and Chen, Huajun},
  journal={arXiv preprint arXiv:2212.09597},
  year={2022}
}

@inproceedings{hoang-etal-2024-toxcl,
    title = "{T}o{XCL}: A Unified Framework for Toxic Speech Detection and Explanation",
    author = "Hoang, Nhat  and
      Do, Xuan Long  and
      Do, Duc Anh  and
      Vu, Duc Anh  and
      Luu, Anh Tuan",
    editor = "Duh, Kevin  and
      Gomez, Helena  and
      Bethard, Steven",
    booktitle = "Proceedings of the 2024 Conference of the North American Chapter of the Association for Computational Linguistics: Human Language Technologies (Volume 1: Long Papers)",
    month = jun,
    year = "2024",
    address = "Mexico City, Mexico",
    publisher = "Association for Computational Linguistics",
    url = "https://aclanthology.org/2024.naacl-long.359",
    doi = "10.18653/v1/2024.naacl-long.359",
    pages = "6460--6472",
}

@article{nguyen2021enriching,
  title={Enriching and controlling global semantics for text summarization},
  author={Nguyen, Thong and Luu, Anh Tuan and Lu, Truc and Quan, Tho},
  journal={arXiv preprint arXiv:2109.10616},
  year={2021}
}

@inproceedings{nguyen2022improving,
  title={Improving neural cross-lingual abstractive summarization via employing optimal transport distance for knowledge distillation},
  author={Nguyen, Thong Thanh and Luu, Anh Tuan},
  booktitle={Proceedings of the AAAI Conference on Artificial Intelligence},
  volume={36},
  number={10},
  pages={11103--11111},
  year={2022}
}

@article{du2023improving,
  title={Improving factuality and reasoning in language models through multiagent debate},
  author={Du, Yilun and Li, Shuang and Torralba, Antonio and Tenenbaum, Joshua B and Mordatch, Igor},
  journal={arXiv preprint arXiv:2305.14325},
  year={2023}
}

@article{liang2023encouraging,
  title={Encouraging divergent thinking in large language models through multi-agent debate},
  author={Liang, Tian and He, Zhiwei and Jiao, Wenxiang and Wang, Xing and Wang, Yan and Wang, Rui and Yang, Yujiu and Shi, Shuming and Tu, Zhaopeng},
  journal={arXiv preprint arXiv:2305.19118},
  year={2023}
}

@article{xiong2023examining,
  title={Examining inter-consistency of large language models collaboration: An in-depth analysis via debate},
  author={Xiong, Kai and Ding, Xiao and Cao, Yixin and Liu, Ting and Qin, Bing},
  journal={arXiv preprint arXiv:2305.11595},
  year={2023}
}

@article{madaan2024self,
  title={Self-refine: Iterative refinement with self-feedback},
  author={Madaan, Aman and Tandon, Niket and Gupta, Prakhar and Hallinan, Skyler and Gao, Luyu and Wiegreffe, Sarah and Alon, Uri and Dziri, Nouha and Prabhumoye, Shrimai and Yang, Yiming and others},
  journal={Advances in Neural Information Processing Systems},
  volume={36},
  year={2024}
}

@article{wang2022self,
  title={Self-consistency improves chain of thought reasoning in language models},
  author={Wang, Xuezhi and Wei, Jason and Schuurmans, Dale and Le, Quoc and Chi, Ed and Narang, Sharan and Chowdhery, Aakanksha and Zhou, Denny},
  journal={arXiv preprint arXiv:2203.11171},
  year={2022}
}

@article{yoran2023answering,
  title={Answering questions by meta-reasoning over multiple chains of thought},
  author={Yoran, Ori and Wolfson, Tomer and Bogin, Ben and Katz, Uri and Deutch, Daniel and Berant, Jonathan},
  journal={arXiv preprint arXiv:2304.13007},
  year={2023}
}

@article{zhou2023thread,
  title={Thread of thought unraveling chaotic contexts},
  author={Zhou, Yucheng and Geng, Xiubo and Shen, Tao and Tao, Chongyang and Long, Guodong and Lou, Jian-Guang and Shen, Jianbing},
  journal={arXiv preprint arXiv:2311.08734},
  year={2023}
}

@article{jin2023tab,
  title={Tab-cot: Zero-shot tabular chain of thought},
  author={Jin, Ziqi and Lu, Wei},
  journal={arXiv preprint arXiv:2305.17812},
  year={2023}
}

@article{khalifa2023exploring,
  title={Exploring demonstration ensembling for in-context learning},
  author={Khalifa, Muhammad and Logeswaran, Lajanugen and Lee, Moontae and Lee, Honglak and Wang, Lu},
  journal={arXiv preprint arXiv:2308.08780},
  year={2023}
}

@article{wang2023rolellm,
  title={Rolellm: Benchmarking, eliciting, and enhancing role-playing abilities of large language models},
  author={Wang, Zekun Moore and Peng, Zhongyuan and Que, Haoran and Liu, Jiaheng and Zhou, Wangchunshu and Wu, Yuhan and Guo, Hongcheng and Gan, Ruitong and Ni, Zehao and Yang, Jian and others},
  journal={arXiv preprint arXiv:2310.00746},
  year={2023}
}

@article{li2023large,
  title={Large language models understand and can be enhanced by emotional stimuli},
  author={Li, Cheng and Wang, Jindong and Zhang, Yixuan and Zhu, Kaijie and Hou, Wenxin and Lian, Jianxun and Luo, Fang and Yang, Qiang and Xie, Xing},
  journal={arXiv preprint arXiv:2307.11760},
  year={2023}
}

@article{lyu2023faithful,
  title={Faithful chain-of-thought reasoning},
  author={Lyu, Qing and Havaldar, Shreya and Stein, Adam and Zhang, Li and Rao, Delip and Wong, Eric and Apidianaki, Marianna and Callison-Burch, Chris},
  journal={arXiv preprint arXiv:2301.13379},
  year={2023}
}

@article{zhou2022least,
  title={Least-to-most prompting enables complex reasoning in large language models},
  author={Zhou, Denny and Sch{\"a}rli, Nathanael and Hou, Le and Wei, Jason and Scales, Nathan and Wang, Xuezhi and Schuurmans, Dale and Cui, Claire and Bousquet, Olivier and Le, Quoc and others},
  journal={arXiv preprint arXiv:2205.10625},
  year={2022}
}

@article{khot2022decomposed,
  title={Decomposed prompting: A modular approach for solving complex tasks},
  author={Khot, Tushar and Trivedi, Harsh and Finlayson, Matthew and Fu, Yao and Richardson, Kyle and Clark, Peter and Sabharwal, Ashish},
  journal={arXiv preprint arXiv:2210.02406},
  year={2022}
}

@article{brown2020language,
  title={Language models are few-shot learners},
  author={Brown, Tom B},
  journal={arXiv preprint arXiv:2005.14165},
  year={2020}
}

@inproceedings{zhao2021calibrate,
  title={Calibrate before use: Improving few-shot performance of language models},
  author={Zhao, Zihao and Wallace, Eric and Feng, Shi and Klein, Dan and Singh, Sameer},
  booktitle={International conference on machine learning},
  pages={12697--12706},
  year={2021},
  organization={PMLR}
}

@article{lum2024bias,
  title={Bias in language models: Beyond trick tests and toward ruted evaluation},
  author={Lum, Kristian and Anthis, Jacy Reese and Nagpal, Chirag and D'Amour, Alexander},
  journal={arXiv preprint arXiv:2402.12649},
  year={2024}
}

@inproceedings{wolf-etal-2020-transformers,
    title = "Transformers: State-of-the-Art Natural Language Processing",
    author = "Wolf, Thomas  and
      Debut, Lysandre  and
      Sanh, Victor  and
      Chaumond, Julien  and
      Delangue, Clement  and
      Moi, Anthony  and
      Cistac, Pierric  and
      Rault, Tim  and
      Louf, Remi  and
      Funtowicz, Morgan  and
      Davison, Joe  and
      Shleifer, Sam  and
      von Platen, Patrick  and
      Ma, Clara  and
      Jernite, Yacine  and
      Plu, Julien  and
      Xu, Canwen  and
      Le Scao, Teven  and
      Gugger, Sylvain  and
      Drame, Mariama  and
      Lhoest, Quentin  and
      Rush, Alexander",
    editor = "Liu, Qun  and
      Schlangen, David",
    booktitle = "Proceedings of the 2020 Conference on Empirical Methods in Natural Language Processing: System Demonstrations",
    month = oct,
    year = "2020",
    address = "Online",
    publisher = "Association for Computational Linguistics",
    url = "https://aclanthology.org/2020.emnlp-demos.6/",
    doi = "10.18653/v1/2020.emnlp-demos.6",
    pages = "38--45"
}

@inproceedings{vu2025curriculum,
  title={Curriculum Demonstration Selection for In-Context Learning},
  author={Vu, Duc Anh and Nguyen, Cong-Duy and Wu, Xiaobao and Hoang, Nhat and Du, Mingzhe and Nguyen, Thong and Luu, Anh Tuan},
  booktitle={Proceedings of the 40th ACM/SIGAPP Symposium on Applied Computing},
  pages={1004--1006},
  year={2025}
}

@inproceedings{zhao-etal-2024-knn,
    title = "$k${NN}-{ICL}: Compositional Task-Oriented Parsing Generalization with Nearest Neighbor In-Context Learning",
    author = "Zhao, Wenting  and
      Liu, Ye  and
      Wan, Yao  and
      Wang, Yibo  and
      Wu, Qingyang  and
      Deng, Zhongfen  and
      Du, Jiangshu  and
      Liu, Shuaiqi  and
      Xu, Yunlong  and
      Yu, Philip",
    editor = "Duh, Kevin  and
      Gomez, Helena  and
      Bethard, Steven",
    booktitle = "Proceedings of the 2024 Conference of the North American Chapter of the Association for Computational Linguistics: Human Language Technologies (Volume 1: Long Papers)",
    month = jun,
    year = "2024",
    address = "Mexico City, Mexico",
    publisher = "Association for Computational Linguistics",
    url = "https://aclanthology.org/2024.naacl-long.19/",
    doi = "10.18653/v1/2024.naacl-long.19",
    pages = "326--337"
}

@inproceedings{chen-etal-2022-improving,
    title = "Improving In-Context Few-Shot Learning via Self-Supervised Training",
    author = "Chen, Mingda  and
      Du, Jingfei  and
      Pasunuru, Ramakanth  and
      Mihaylov, Todor  and
      Iyer, Srini  and
      Stoyanov, Veselin  and
      Kozareva, Zornitsa",
    editor = "Carpuat, Marine  and
      de Marneffe, Marie-Catherine  and
      Meza Ruiz, Ivan Vladimir",
    booktitle = "Proceedings of the 2022 Conference of the North American Chapter of the Association for Computational Linguistics: Human Language Technologies",
    month = jul,
    year = "2022",
    address = "Seattle, United States",
    publisher = "Association for Computational Linguistics",
    url = "https://aclanthology.org/2022.naacl-main.260/",
    doi = "10.18653/v1/2022.naacl-main.260",
    pages = "3558--3573"
}

@inproceedings{nguyen-etal-2023-improving-multimodal,
    title = "Improving Multimodal Sentiment Analysis: Supervised Angular margin-based Contrastive Learning for Enhanced Fusion Representation",
    author = "Nguyen, Cong-Duy  and
      Nguyen, Thong  and
      Vu, Duc  and
      Luu, Anh",
    editor = "Bouamor, Houda  and
      Pino, Juan  and
      Bali, Kalika",
    booktitle = "Findings of the Association for Computational Linguistics: EMNLP 2023",
    month = dec,
    year = "2023",
    address = "Singapore",
    publisher = "Association for Computational Linguistics",
    url = "https://aclanthology.org/2023.findings-emnlp.980/",
    doi = "10.18653/v1/2023.findings-emnlp.980",
    pages = "14714--14724"
}

@article{nguyen2025cutpaste,
  title={CutPaste\&Find: Efficient Multimodal Hallucination Detector with Visual-aid Knowledge Base},
  author={Nguyen, Cong-Duy and Wu, Xiaobao and Vu, Duc Anh and Zhao, Shuai and Nguyen, Thong and Luu, Anh Tuan},
  journal={arXiv preprint arXiv:2502.12591},
  year={2025}
}

@article{liu2023visual,
  title={Visual instruction tuning},
  author={Liu, Haotian and Li, Chunyuan and Wu, Qingyang and Lee, Yong Jae},
  journal={Advances in neural information processing systems},
  volume={36},
  pages={34892--34916},
  year={2023}
}

@inproceedings{pezeshkpour-hruschka-2024-large,
    title = "Large Language Models Sensitivity to The Order of Options in Multiple-Choice Questions",
    author = "Pezeshkpour, Pouya  and
      Hruschka, Estevam",
    editor = "Duh, Kevin  and
      Gomez, Helena  and
      Bethard, Steven",
    booktitle = "Findings of the Association for Computational Linguistics: NAACL 2024",
    month = jun,
    year = "2024",
    address = "Mexico City, Mexico",
    publisher = "Association for Computational Linguistics",
    url = "https://aclanthology.org/2024.findings-naacl.130/",
    doi = "10.18653/v1/2024.findings-naacl.130",
    pages = "2006--2017"
}

@inproceedings{liu2024we,
  title={" we need structured output": Towards user-centered constraints on large language model output},
  author={Liu, Michael Xieyang and Liu, Frederick and Fiannaca, Alexander J and Koo, Terry and Dixon, Lucas and Terry, Michael and Cai, Carrie J},
  booktitle={Extended Abstracts of the CHI Conference on Human Factors in Computing Systems},
  pages={1--9},
  year={2024}
}

@misc{chen2022improvingincontextfewshotlearning,
      title={Improving In-Context Few-Shot Learning via Self-Supervised Training}, 
      author={Mingda Chen and Jingfei Du and Ramakanth Pasunuru and Todor Mihaylov and Srini Iyer and Veselin Stoyanov and Zornitsa Kozareva},
      year={2022},
      eprint={2205.01703},
      archivePrefix={arXiv},
      primaryClass={cs.CL},
      url={https://arxiv.org/abs/2205.01703}, 
}

@misc{khot2023decomposedpromptingmodularapproach,
      title={Decomposed Prompting: A Modular Approach for Solving Complex Tasks}, 
      author={Tushar Khot and Harsh Trivedi and Matthew Finlayson and Yao Fu and Kyle Richardson and Peter Clark and Ashish Sabharwal},
      year={2023},
      eprint={2210.02406},
      archivePrefix={arXiv},
      primaryClass={cs.CL},
      url={https://arxiv.org/abs/2210.02406}, 
}

@misc{khalifa2023exploringdemonstrationensemblingincontext,
      title={Exploring Demonstration Ensembling for In-context Learning}, 
      author={Muhammad Khalifa and Lajanugen Logeswaran and Moontae Lee and Honglak Lee and Lu Wang},
      year={2023},
      eprint={2308.08780},
      archivePrefix={arXiv},
      primaryClass={cs.CL},
      url={https://arxiv.org/abs/2308.08780}, 
}

@misc{vu2024curriculumdemonstrationselectionincontext,
      title={Curriculum Demonstration Selection for In-Context Learning}, 
      author={Duc Anh Vu and Nguyen Tran Cong Duy and Xiaobao Wu and Hoang Minh Nhat and Du Mingzhe and Nguyen Thanh Thong and Anh Tuan Luu},
      year={2024},
      eprint={2411.18126},
      archivePrefix={arXiv},
      primaryClass={cs.CL},
      url={https://arxiv.org/abs/2411.18126}, 
}

@misc{medical,
author = {Klang, Eyal and Portugez, Shir and Gross, Raz and Kassif, Reut and Brenner, Alina and Gilboa, Maayan and Ortal, Tal and Ron, Sophi and Robinzon, Vered and Meiri, Hila and Segal, Gadi},
year = {2023},
month = {07},
pages = {},
title = {Utilizing Artificial Intelligence for Crafting Medical Examinations: A Medical Education Study with GPT-4},
doi = {10.21203/rs.3.rs-3146947/v1}
}

@misc{lum2025biaslanguagemodelstrick,
      title={Bias in Language Models: Beyond Trick Tests and Toward RUTEd Evaluation}, 
      author={Kristian Lum and Jacy Reese Anthis and Kevin Robinson and Chirag Nagpal and Alexander D'Amour},
      year={2025},
      eprint={2402.12649},
      archivePrefix={arXiv},
      primaryClass={cs.CL},
      url={https://arxiv.org/abs/2402.12649}, 
}

@misc{wang2024llmsperformmcqaselecting,
      title={LLMs May Perform MCQA by Selecting the Least Incorrect Option}, 
      author={Haochun Wang and Sendong Zhao and Zewen Qiang and Nuwa Xi and Bing Qin and Ting Liu},
      year={2024},
      eprint={2402.01349},
      archivePrefix={arXiv},
      primaryClass={cs.CL},
      url={https://arxiv.org/abs/2402.01349}, 
}

@article{tomova2024leveraging,
  title={Leveraging large language models to construct feedback from medical multiple-choice questions},
  author={Tomova, Mihaela and Rosell{\'o} Atanet, Iv{\'a}n and Sehy, Victoria and Sieg, Miriam and M{\"a}rz, Maren and M{\"a}der, Patrick},
  journal={Scientific reports},
  volume={14},
  number={1},
  pages={27910},
  year={2024},
  publisher={Nature Publishing Group UK London}
}

@article{wu2024survey,
  title={A survey on neural topic models: methods, applications, and challenges},
  author={Wu, Xiaobao and Nguyen, Thong and Luu, Anh Tuan},
  journal={Artificial Intelligence Review},
  volume={57},
  number={2},
  pages={18},
  year={2024},
  publisher={Springer Netherlands Dordrecht}
}

@inproceedings{nguyen2024video,
  title={Video-Language Understanding: A Survey from Model Architecture, Model Training, and Data Perspectives},
  author={Nguyen, Thong and Bin, Yi and Xiao, Junbin and Qu, Leigang and Li, Yicong and Wu, Jay Zhangjie and Nguyen, Cong-Duy and Ng, See-Kiong and Tuan, Luu Anh},
  booktitle={ACL 2024 (Findings)},
  year={2024}
}

@article{nguyen2024meta,
  title={Meta-optimized Angular Margin Contrastive Framework for Video-Language Representation Learning},
  author={Nguyen, Thong and Bin, Yi and Wu, Xiaobao and Dong, Xinshuai and Hu, Zhiyuan and Le, Khoi and Nguyen, Cong-Duy and Ng, See-Kiong and Tuan, Luu Anh},
  journal={ECCV 2024},
  year={2024}
}

@inproceedings{nguyen2024encoding,
  title={Encoding and controlling global semantics for long-form video question answering},
  author={Nguyen, Thong Thanh and Hu, Zhiyuan and Wu, Xiaobao and Nguyen, Cong-Duy T and Ng, See Kiong and Tuan, Luu Anh},
  booktitle={Proceedings of the 2024 Conference on Empirical Methods in Natural Language Processing},
  pages={7049--7066},
  year={2024}
}

@article{wu2024fastopic,
  title={Fastopic: Pretrained transformer is a fast, adaptive, stable, and transferable topic model},
  author={Wu, Xiaobao and Nguyen, Thong and Zhang, Delvin and Wang, William Yang and Luu, Anh Tuan},
  journal={Advances in Neural Information Processing Systems},
  volume={37},
  pages={84447--84481},
  year={2024}
}

@inproceedings{nguyen2025motion,
  title={Motion-aware contrastive learning for temporal panoptic scene graph generation},
  author={Nguyen, Thong Thanh and Wu, Xiaobao and Bin, Yi and Nguyen, Cong-Duy T and Ng, See-Kiong and Luu, Anh Tuan},
  booktitle={Proceedings of the AAAI Conference on Artificial Intelligence},
  volume={39},
  number={6},
  pages={6218--6226},
  year={2025}
}


% \section{Additional Analysis on BiasPrompting Performance}

% \section{Average length of generated reasonings}

% \begin{figure*}[!]
    
%     \centering
%     \includegraphics[width=\textwidth]{images/avg_length.pdf}
%     \caption{Violin plots comparing the average response lengths generated by BiasPrompting and CoT prompting across five datasets (CSQA, SQA, PIQA, DU, and CJ). The {\color[HTML]{86b66c} $\blacksquare$} violins represent BiasPrompting, while the {\color[HTML]{78acd1} $\blacksquare$} violins correspond to CoT prompting. The dashed line inside each violin indicates the mean response length.}
%     \label{fig:avglength}
% \end{figure*}

% \section{Additional Experimental Details}
% \label{additional}

% We use the weights of three 7B parameter models, Mistral-7B\cite{jiang2023mistral}\footnote{\url{https://huggingface.co/mistralai/Mistral-7B-Instruct-v0.2}}, Deepseek-7B\cite{bi2024deepseek}\footnote{\url{https://huggingface.co/deepseek-ai/deepseek-llm-7b-chat}}, and Gemma-7B\cite{Mesnard2024GemmaOM}\footnote{\url{https://huggingface.co/google/gemma-7b-it}}, sourced from Huggingface. 

\end{document}